\documentclass[sigconf]{acmart}

\usepackage{xspace}
\usepackage{multirow}
\usepackage{balance}

\AtBeginDocument{%
  \providecommand\BibTeX{{%
    \normalfont B\kern-0.5em{\scshape i\kern-0.25em b}\kern-0.8em\TeX}}}

\copyrightyear{2021}
\acmYear{2021}
\setcopyright{acmcopyright}\acmConference[MM '21]{Proceedings of the 29th ACM International Conference on Multimedia}{October 20--24, 2021}{Virtual Event, China}
\acmBooktitle{Proceedings of the 29th ACM International Conference on Multimedia (MM '21), October 20--24, 2021, Virtual Event, China}
\acmPrice{15.00}
\acmDOI{10.1145/3474085.3475206}
\acmISBN{978-1-4503-8651-7/21/10}

\begin{document}
\fancyhead{} 
\title{Towards Controllable and Photorealistic \\ Region-wise Image Manipulation}



\author{Ansheng You}
\authornote{These authors contributed equally to this work and share the first authorship.}
\email{youansheng@pku.edu.cn}
\affiliation{
  \institution{Peking University}
  \city{Beijing}
  \country{China}
}

\author{Chenglin Zhou}
\email{zhouchl@shanghaitech.edu.cn}
\authornotemark[1]
\affiliation{
  \institution{ShanghaiTech University}
  \city{Shanghai}
  \country{China}
}

\author{Qixuan Zhang}
\email{zhangqx1@shanghaitech.edu.cn}
\affiliation{
  \institution{ShanghaiTech University}
  \city{Shanghai}
  \country{China}
}

\author{Lan Xu} 
\email{xulan1@shanghaitech.edu.cn}
\affiliation{
  \institution{ShanghaiTech University}
  \city{Shanghai}
  \country{China}
}

\renewcommand{\shortauthors}{Ansheng You, et al.}

\makeatletter
\DeclareRobustCommand\onedot{\futurelet\@let@token\@onedot}
\def\@onedot{\ifx\@let@token.\else.\null\fi\xspace}

\def\eg{\emph{e.g}\onedot} \def\Eg{\emph{E.g}\onedot}
\def\ie{\emph{i.e}\onedot} \def\Ie{\emph{I.e}\onedot}
\def\cf{\emph{c.f}\onedot} \def\Cf{\emph{C.f}\onedot}
\def\etc{\emph{etc}\onedot} \def\vs{\emph{vs}\onedot}
\def\wrt{w.r.t\onedot} \def\dof{d.o.f\onedot}
\def\etal{\emph{et al}\onedot}
\makeatother

\begin{abstract}
   Adaptive and flexible image editing is a desirable function of modern generative models. In this work, we present a generative model with auto-encoder architecture for per-region style manipulation. We apply a code consistency loss to enforce an explicit disentanglement between content and style latent representations, making the content and style of generated samples consistent with their corresponding content and style references. The model is also constrained by a content alignment loss to ensure the foreground editing will not interfere background contents. As a result, given interested region masks provided by users, our model supports foreground region-wise style transfer. Specially, our model receives no extra annotations such as semantic labels except for self-supervision. Extensive experiments show the effectiveness of the proposed method and exhibit the flexibility of the proposed model for various applications, including region-wise style editing, latent space interpolation, cross-domain style transfer. 
\end{abstract}

\ccsdesc[500]{Computing methodologies~Computer vision}
\ccsdesc[300]{Computing methodologies~Computer vision problems}
\ccsdesc[300]{Computing methodologies~Reconstruction}
\ccsdesc[300]{Computing methodologies~Appearance and texture representations}
\keywords{image editing; controllable and photorealistic; code consistency; content alignment}

\begin{teaserfigure}
		\centering
		\includegraphics[width=\textwidth]{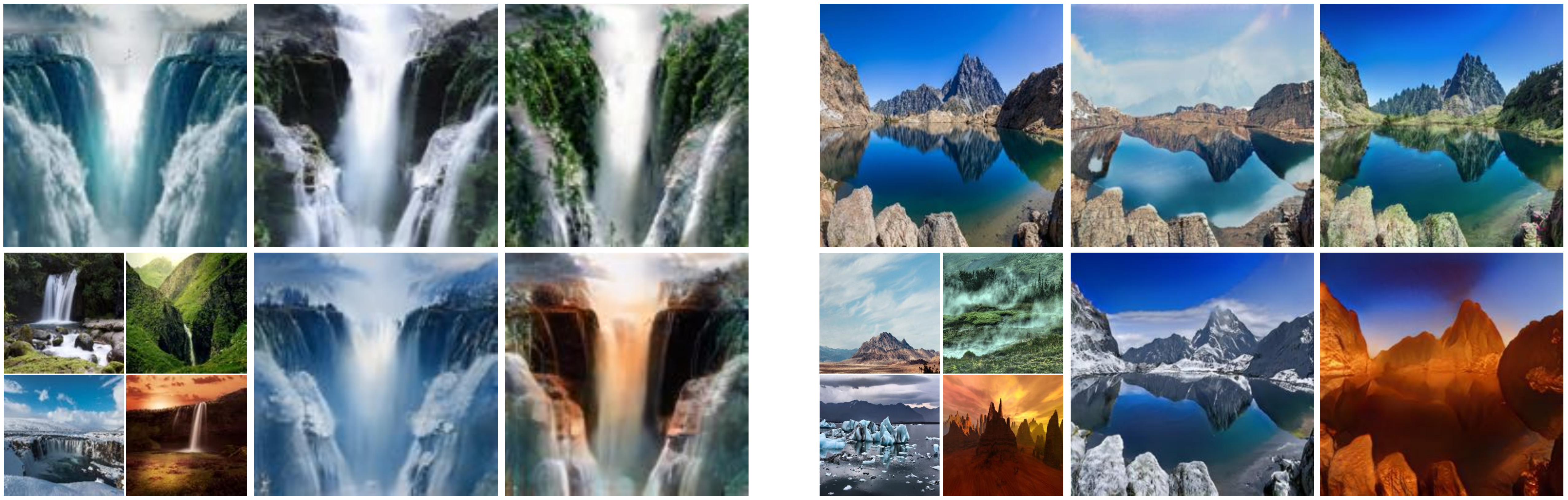}
	\captionof{figure}{Photorealistic image style transfer by our method on waterfall and mountain. The top-left image of each image block represents the content reference image and the bottom-left four images are the style reference images. Our synthesized results are placed orderly on the right following the layout of the style reference images. The reference images are taken from \url{https://pixabay.com/}. \textbf{Zoom in} for better details.
			\label{fig:teaser}
		}
\end{teaserfigure}

\maketitle

\section{Introduction}

Image editing has numerous applications in real-world life, such as artistic content creation, make-up transfer, virtual reality, gaming, \emph{etc.}
At the basic level, image editing is about either altering the image contents (objects and their layout in the image) or changing the styles or textures of the image. 
In this paper, we focus on the latter one and aim to design a new algorithm that fulfills the needs of users to freely and conveniently edit the styles of {\em real} images. Some of our generated samples are shown in Fig.~\ref{fig:teaser}.

Essentially, style transfer is to extract the contents from the current image and attach style statistics obtained from other images to the current image. We assume an image is mainly composed by a content code and style code. Therefore, the disentanglement between these two codes is crucial to the performance: too relaxed form will lead to entangled representation, and make the result either be too similar to the style image or the content image. The superior performance of \cite{SwapAuto} shows a basic architectural design difference for the content code and style code is possible to perform a \emph{soft} code disentanglement. However, we notice with training, it is hard to balance the disentanglement and the generated results tend to lose structure consistency with their original structures. Considering this problem, we propose to enforce a {\bf code consistency loss} based on self-supervision to \emph{explicitly} disentangle content and style codes. Specifically, we obtain the content and style codes of the transferred image and then enforce its content code similar to the original content code, and its style code equal to the style code of the style reference. With this constraint, we observe an obvious improvement over \cite{SwapAuto} in keeping the structure of the generated image to the content reference and style similar to the style reference.

Inspired by the works~\cite{SPADE,SEAN,SMIS} in label-to-image translation task, we believe more flexible user control benefits from region-wise image style transfer in our scenario. Users can freely change the style of selected areas in the image while keep other parts maintaining their original styles. Considering the cost and constraint to use semantic labels~\cite{SMIS,SEAN}, we are leaning towards unconditional approaches without using semantic labels. The essential requirement of region-wise stylization is to change the style of the given area while maintaining the style of other area in the image. Consequently, we use a randomly generated mask to differentiate foreground and background to represent areas to be stylized or not. Naively, the background is supposed to keep static to the original image but the target for foreground area is
\emph{unknown}. Hinted by the above self-supervised training used to conduct code disentanglement, we can leverage the transferred image as the \emph{pseudo} ground-truth for the foreground area. In this way, we propose a {\bf content alignment loss} to constrain the foreground area to be similar to that of the transferred image and the background maintaining unchanged. Such loss function enables our model to perform region-wise stylization, without the use of expensive semantic label annotations.

To summarize, we propose two loss functions, namely the code consistency loss and the content alignment loss, to explicitly disentangle content and style codes for better performance and empowers the generator with the ability of region-wise stylization, respectively. Our model, named {\bf TCP}, is only trained with self-supervison without other annotations. Yet, it can finely edit real image styles and shows superior performance on various datasets, such as FFHQ~\cite{StyleGAN}, LSUN-Church~\cite{LSUN}, and our collected waterfall and mountain datasets as~\cite{SwapAuto}. Experimental results also show the prospect of applications like content or style code interpolation and cross-domain style transfer.

\section{Related Works}

\vspace{0.5ex}\noindent \textbf{Conditional image generation.}~Many applications aim to learn an image mapping from the conditional image inputs to desired image domains. The advent of generative adversarial networks~\cite{GAN,CGAN} had driven a thread of useful conditional image generation applications, such as general image-to-image translation~\cite{pix2pix,MUNIT,pix2pixHD,CycleGAN,DIRT,UNIT,StarGAN,StarGAN2}, image inpainting~\cite{PartialConv,FreeForm,ContextualAttention}, style transfer~\cite{gatys1,Gatys2016,CBN,AdaIN,PhotoWCT,DPST,WCT,WCT2}, texture synthesis~\cite{NonStaionary,TextureGAN,DiversifiedTexture}, semantic label to image~\cite{SPADE,pix2pixHD,SEAN,SMIS}, \etc. 

\vspace{0.5ex}\noindent \textbf{Image manipulation through latent representation.}~Recently, researchers are trying to generate high-resolution images via designing new generator architectures~\cite{StyleGAN,STYLEGAN2,pix2pixHD}, training policies~\cite{ProgressiveGAN,BigGAN}, single image learning strategies~\cite{NonStaionary,SinGAN,InGAN}, \etc. Among them, StyleGAN~\cite{StyleGAN} and StyleGAN2~\cite{STYLEGAN2} proposed a style modulating generator which learns from random noises of Gaussian. Such generators have very smooth latent space that can be injected to multiple resolution scales as style modulation parameters. These parameters can control different aspects of the generated images. There are several representative attempts aiming to gain more flexible and specific control upon the generation process by identifying editable latent or feature directions: some~\cite{GANSpace} performed Principal Components Analysis to the latent space or feature space for direction identification; Alharbi~\etal~\cite{NoiseInjection} injected noises to the generator to obtain spatial control; Shen~\etal~\cite{InterFaceGAN,InterpretFaceEditing} used face attribute classifiers to assist the decouple of face attributes for face manipulation; Shen~\etal~\cite{ClosedFormFactorization} also presented a closed-form factorization method to discover the semantic attributes learned in network parameters; Peebles~\etal~\cite{HessianPenalty} introduced a regularization term to encourage the generator to be more disentangled by minimizing the off-diagonal entries of the hessian matrix of the generator. 

\vspace{0.5ex}\noindent \textbf{Real image manipulation.}~Though image editing through latent space manipulation is extensively studied recently, performing editing operations on real images is non-trivial. It requires a must-step to encode the real images to editable codes. Some works~\cite{STYLEGAN2,Im2StyleGAN,Im2StyleGAN2,Invertibility} tried to solve this problem by fixing the learned generator and optimizing the latent inputs through minimizing the content losses between the real image and the generated image. This method is computationally expensive, which might needs minutes of optimization for a single image. Another more efficient line of works~\cite{SemanticManipulation,GenerativeVisual,SwapAuto} added an extra encoder ahead of the generator to map real image inputs to latent codes. Such method is much more efficient except that it needs to train an extra encoder. In our work, we also adopt the encoder-based mapping for better efficiency and we use the encoder as a requisite to compute our losses.


\vspace{0.5ex}\noindent \textbf{Region-wise control.}~Region-wise editing is a commonly used operations of user-interactive content creation system. Some works like SEAN~\cite{SEAN} and SMIS~\cite{SMIS} made architectural change to SPADE~\cite{SPADE} to enable style manipulation over specific semantic area. Though they have much larger flexibility and stronger controllability over semantic region control, they need costly semantic annotations in both training and inference. Different from them, this work shows through our self-supervised training strategy, our model can also perform some extent of region-wise style editing without using semantic labels for training. 

\vspace{0.5ex}\noindent \textbf{Style transfer.} Except for structural modification, style transfer can be regarded as another mainstream image manipulation. Style transfer aims to swap low-level statistics of an image, such as color, illumination, textures, with those from another image. Current works of this topic leveraged deep neural networks~\cite{Gatys2016,gatys1,perceptualloss,MarkovianGAN} by minimizing the style statistics, \eg Gram matrix, between generated and style reference images and enforcing pixel-wise content losses to ensure content consistency. These works support local style changes but have problems to cope with larger structure change. Usually, the results expose artistic pattern. Photo-realistic style transfer methods~\cite{PhotoWCT,DPST,WCT2} intend to keep results more rigid to inputs and mainly conduct local color transformation. Similar to SwapAuto~\cite{SwapAuto}, we try to propose a model equipped with local structural change capacity and in the meantime, generating photorealistic images with similar styles to the style references.
\section{Method}

\begin{figure}[ht]
	\centering
	\includegraphics[width=0.45\textwidth]{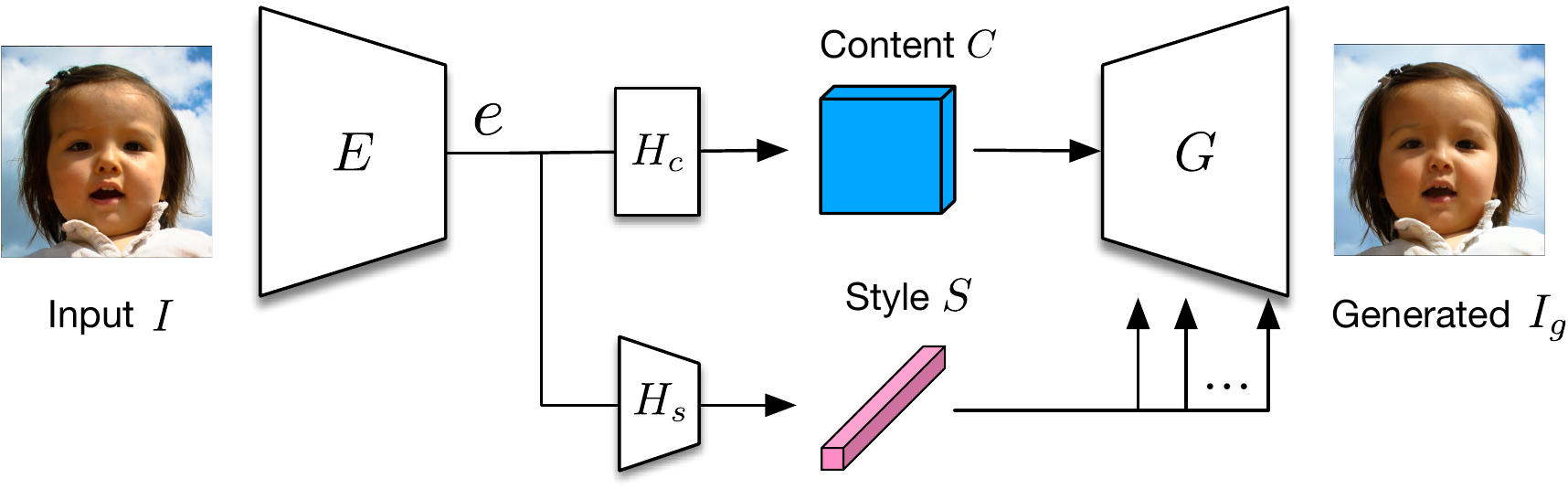}
	\caption{The generation pipeline of our method. 
	\label{fig:model}
	}
\end{figure}
\subsection{Method overview}
\label{sec:method_overview}
Real image editing can be roughly divided into two categories: structure manipulation and image stylization from another reference. This work mainly centers on more controllable and photorealistic image stylization. From the image stylization perspective, any image $I$ is composed by its content $C$ and style $S$. However, it is hard to give a clear and deterministic definition for $C$ and $S$, \eg, the direction of human hair could be regarded as content or otherwise, local style. Nonetheless, in our paper, we wish $C$ to give information about objects, their structures, locations, local textures, \etc, while $S$ to depict low-frequency statistics such as the color of different objects, illumination, \etc.

A brief sketch of our model is depicted in Fig.~\ref{fig:model}. Our main architecture is inspired from~\cite{SwapAuto} that it has an encoder $E$ to project an image input $I$ into deep feature map $e$ and then two sub-networks $H_c$ and $H_s$ to convert $e$ into a content code $C$ and style code $S$, respectively. Note, $H_c$ contains several normal convolutions and then compress the channel dimensions of $C$ to a reasonable number. But $H_s$ is composed by a series of down-sampling convolutions and finally vectorizes $S$ to intentionally strip off structural information. In this way, we wish $S$ to focus on feature statistics about styles, color, illumination, \etc. The generator $G$ follows the architecture of StyleGAN2 generator, whose constant input is changed to $C$ while the style modulation parameters are replaced by $S$. The generator outputs an image $I_g=G(C,S)$ given a random combination of $C$ and $S$.

The generator is trained to perform two tasks: image reconstruction and image style transfer. Suppose two images $I_A$ and $I_B$ are in the dataset. For the reconstruction of $I_A$, the generator performs: $I_{A \rightarrow A}=G(C_A, S_A)$.
{Style transfer from $I_A$ to $I_B$ is about generating an image with its style code from $I_B$ and its content code remaining as $C_A$. }Therefore, the stylized generated image can be written as: $I_{A \rightarrow B}=G(C_A, S_B)$.
To train the reconstruction task, we follow~\cite{SwapAuto} to enforce a $\mathcal{L}_1$ loss between $I_{A \rightarrow A}$ and $I_A$: 
\begin{equation}
    \mathcal{L}_{\mathrm{rec}} = \mathcal{L}_1 (I_{A \rightarrow A}, I_A).
\end{equation} However, the transferred result $I_{A \rightarrow B}$ has no clear target. To facilitate the training, SwapAuto~\cite{SwapAuto} introduced a co-occurrence discriminator $D_{\mathrm{co}}$ to judge whether randomly cropped image patches from $I_{A \rightarrow B}$ have the same marginal distributions as reference patches that are randomly cropped from $I_B$. Additionally, there is another discriminator $D$ applied to judge the fidelity of both $I_{A \rightarrow A}$ and $I_{A \rightarrow B}$. Together, the total adversarial loss can be written as:
\begin{small}
\begin{equation}
\begin{aligned}
\mathcal{L}_{\mathrm{adv}} & = \mathbb{E}_D[-\log(D(I_{A\rightarrow A}))] + \mathbb{E}_D[-\log(D(I_{A\rightarrow B}))] \\
& + \mathbb{E}_{D_{\mathrm{co}}}[-\log(D_{\mathrm{co}}(\mathrm{crop}(I_{A\rightarrow B}), \mathrm{crop}(I_B)))].
\end{aligned}
\end{equation}
\end{small} Under ideal convergence, these two discriminators help to ensure the realism of generated images and the reliable random combination of content and style representation to create a hybrid image\footnote{We use hybrid image to represent the generated image whose style code and content code are not from the same source.}. More training and architecture details can be found in \cite{SwapAuto} and our supplementary materials.

\subsection{Code consistency loss for code disentanglement}

\begin{figure}[ht]
	\centering
	\includegraphics[width=0.45\textwidth]{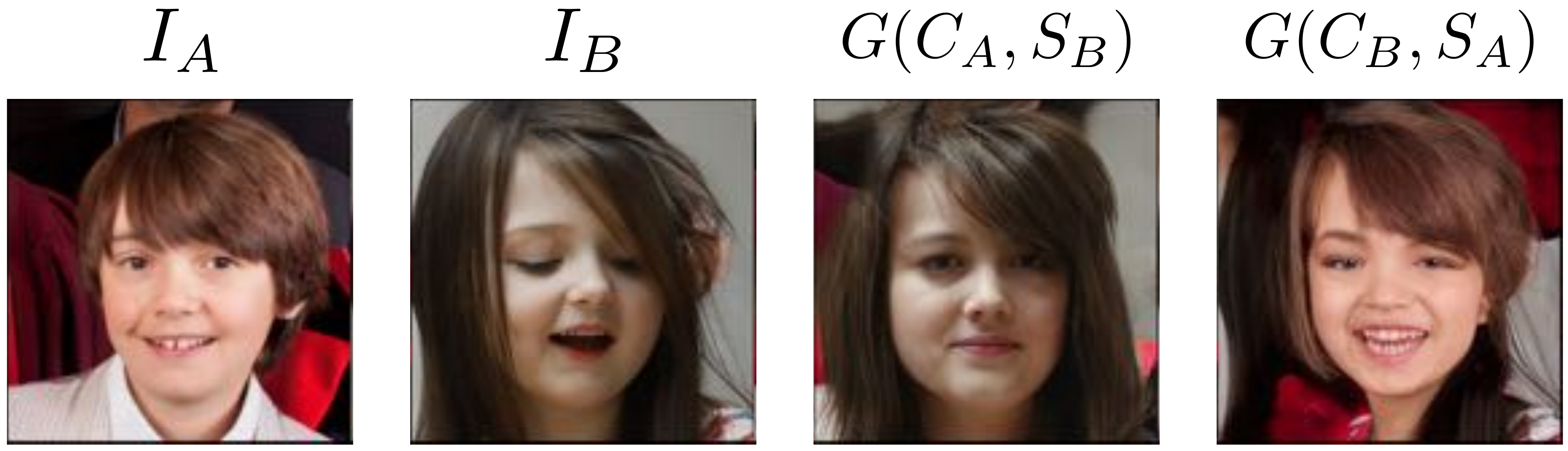}
	\caption{The above results show that if the content code and style code are not properly disentangled, the transferred images are leaning towards the style references, where ideally they should structurally be consistent with their content references.
	\label{fig:consistency_necessity}
	}
\end{figure}

\begin{figure*}[ht]
	\centering
	\includegraphics[width=\textwidth]{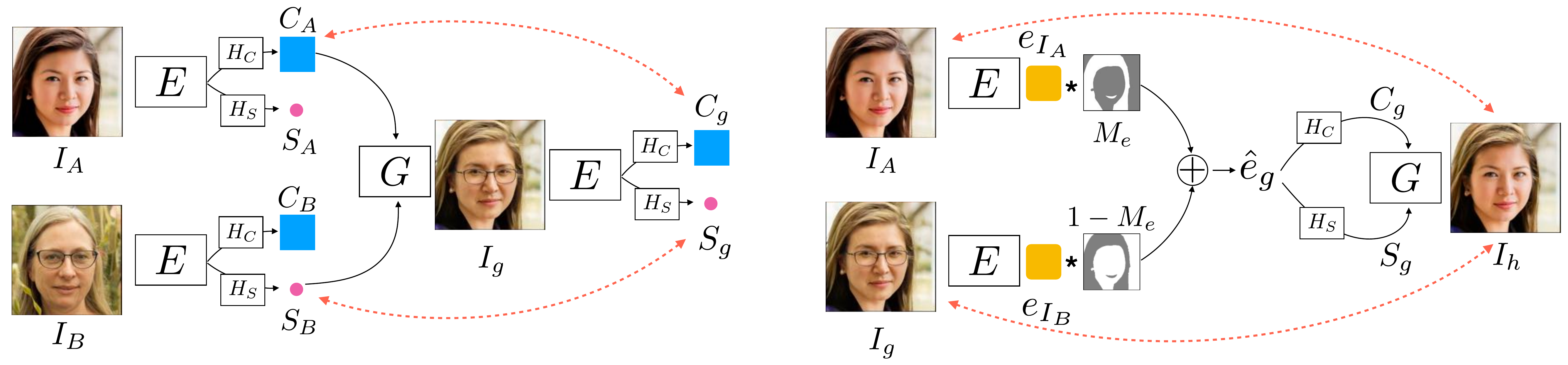}
	\caption{{\bf (Left)} The pipeline to calculate code consistency loss $\mathcal{L}_{\mathrm{CC}}$ after generating globally transferred image $I_g$. {\bf (Right)} The pipeline to calculate the content alignment loss $\mathcal{L}_{\mathrm{CA}}$. Blue squares in the figure denote the content codes; pink dots denote the style codes; the yellow rounded squares represent the output deep features of the encoder. Besides, black solid arrows represent the information flow while red dashed double-sided arrows represent the loss constraints. \textbf{Zoom in} for better details.
	\label{fig:self_supervision}
	}
\end{figure*}

A better disentanglement between the content code and style code is always desirable so that the generator can produce plausible results given any combination of content codes and style codes. Since it is undetermined to differentiate content codes and style codes with clear mathematical formulation, SwapAuto~\cite{SwapAuto} presents a flexible and relaxed form of disentanglement that vectorizes its style code to intentionally lose spatial information about structures (See Sec.~\ref{sec:method_overview}). However, through experiments, we found the results are likely to change the image structure towards the style reference using its default setting, as illustrated in Fig.~\ref{fig:consistency_necessity}. We hypothesize this phenomenon occurs because the structure information is also embedded inside the style code and the generator relies on more the style code to produce the results. This problem motivates us to devise a more explicit code disentanglement strategy, which at least can make the results structurally stay rigid to the content references.

There are more explicit forms of disentanglement such as feature whitening~\cite{WCT,FeatureWhitening} on the content code. However, we experimentally find this fails for our case because the whitening operation is likely to wipe off important structure details. Since our model is an auto-encoder rather than U-Net~\cite{UNet,pix2pix} network, the lost details of structure cannot get supplemented as~\cite{WCT,WCT2} and the model produces highly stylish results without local texture details. 

{We hypothesize that if the encoder is under good convergence and is robust to either real or fake inputs, it can make a fine prediction on the content and style codes of generated image.} Considering the generated image should keep rigid to the content references in structure and consistent to the styles given by the style references, its content code should be similar to that of the content reference, and its style code should be close to that of the style reference. 
Specifically, we feed the globally style transferred result $I_g = G(C_A, S_B)$ to the encoder and the subsequent sub-networks to obtain its content code $C_g$ and style code $S_g$. Ideally, $C_g$ should be equal to $C_A$ and $S_g$ should be close to $S_B$. To ensure this, we introduce a {\bf C}ode {\bf C}onsistency loss $\mathcal{L}_{\mathrm{CC}}$ to enforce these two constraints by using $\mathcal{L}_1$ norm:
\begin{equation}
    \label{eq_self_supervision}
    \mathcal{L}_{\mathrm{CC}} = \mathcal{L}_1 (C_g, C_A) + \lambda \mathcal{L}_1 (S_g, S_B).
\end{equation} The first term is to evaluate the content consistency while the second term reflects the style consistency. $\lambda$ is loss weighting parameter to balance the overall loss scale. 
The above loss function is reminiscent of the Conditional Latent Regressor GAN model as introduced in~\cite{BiCycleGAN}, however, our constraint is performed on real projected codes rather than those randomly drawn from a known Gaussian distribution.

Also note, we made an assumption that the encoder $E$ should be robust to both fake and real inputs. To implement that, we should first train the encoder to optima. However, in practical, we simplify this by alternatively enforcing the losses (Eq.~\ref{eq_self_supervision}) with an interval of $K$ iterations, and we also observe a good state of convergence. In default, $K=16$.

\subsection{Content alignment loss for region-wise stylization}

Our target is to perform region-wise style transfer, which requires an additional input to represent the interested area to change its style. Therefore, we introduce a mask $M$ which has the same spatial size with $I_A$ and $I_B$ to represent the user interested area.  
If using $\{(i,j), 0\leq i<H, 0\leq j<W\}$ to enumerate the pixels in the image space $I\in \mathcal{R}^{H \times W}$, when $M^{(i,j)}=1$, the target hybrid image $I_h$ of region-wise style transfer should show similar styles with $I_B$; for pixels where $M_{(i,j)}=0$, $I_h$ should remain the same with $I_A$.

For region-wise style transfer, the main challenge is to solve when $M^{(i,j)}=1$, what $I_g$ will be look like, considering the result for $M_{(i,j)}=0$ is known. One intuitive solution is to use a pre-trained model $G$ that is able to perform global style transfer to generate a globally stylized image $I_g=G(C_A, S_B)$\footnote{In the following, we use $I_g$ to abbreviate $I_g=G(C_A, S_B)$.} given the input of content reference $I_A$ and style reference $I_B$. The next is to perform pixel-wise combination between $I_g(C_A;S_B)$ and $I_A$:
\begin{equation}
\label{eq1}
I_h^{(i,j)} =\left\{
\begin{aligned}
& I_g^{(i,j)}, & M^{(i,j)}=1, \\
& I_A^{(i,j)}, & M^{(i,j)}=0.
\end{aligned}
\right.
\end{equation}

Though this method is quite reasonable for the task, we find this direct pixel-wise combination leads to severe boundary discordance. 
Around the borders, the pixels from these two images usually show different color distributions. 
However, we assume performing combination in the deep feature space rather than image space is better, as the subsequent networks after such deep features could benefit from large-scale data learning so as to learn the removal of the boundary misalignment to some extent~\cite{DeepImagePrior}.

\begin{figure*}[htbp]
	\centering
	\includegraphics[width=\textwidth]{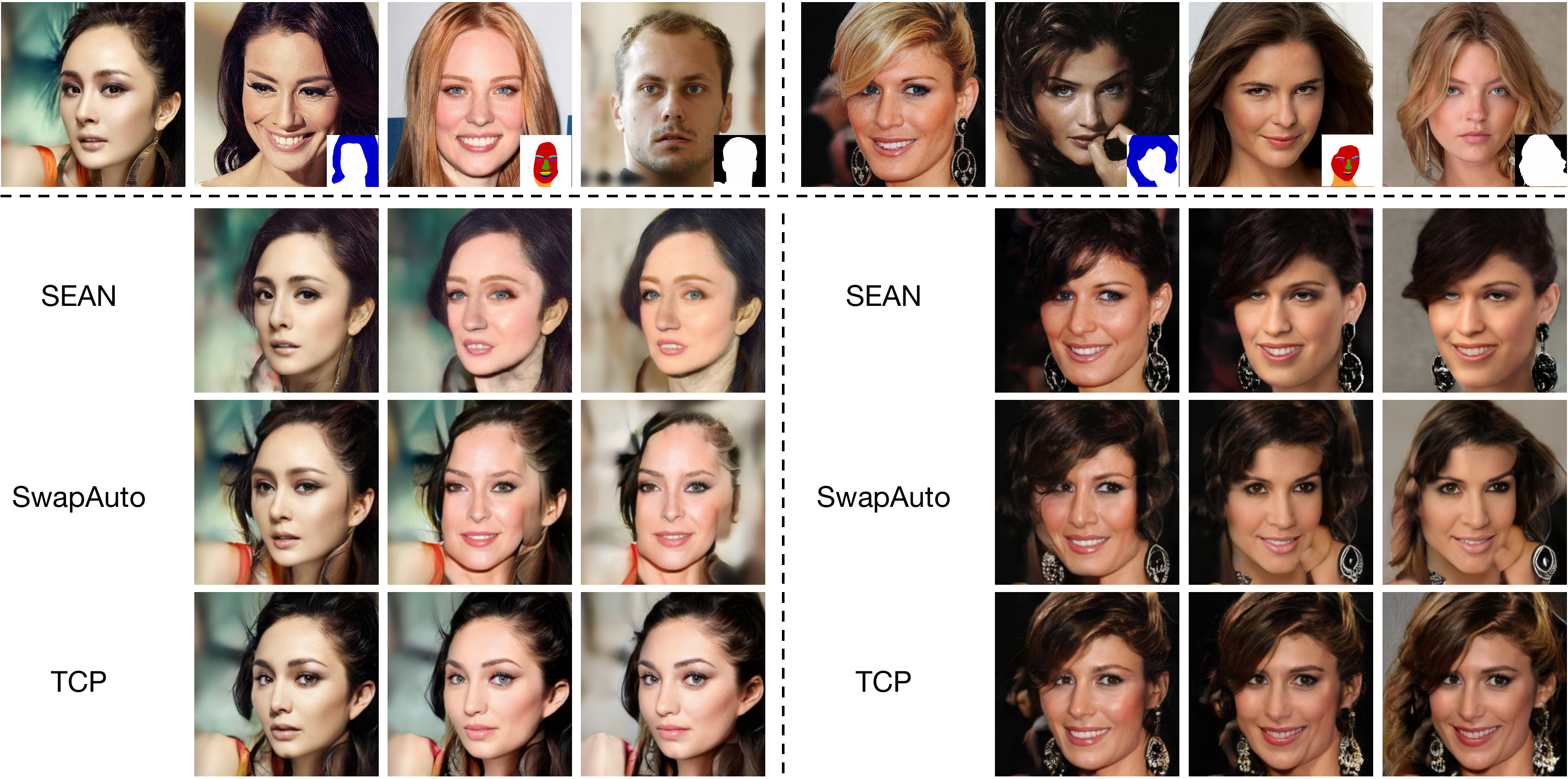}
	\caption{Qualitative comparison of region-wise stylization on the FFHQ dataset. We show two examples separated by vertical dashed line. The top-left of each example is the source image, followed by three style reference images with mask regions of interest in the first row. The results of the successive edits for different approaches are shown in the remaining rows. We use {\bf TCP} to denote our result. 
	\label{fig:comparison_on_regionwise_stylization}
	}
\end{figure*}

To implement our idea, we feed $I_g$ to the encoder $E$ to extract a deep feature map $e_{I_g}$ before going to the consequent sub-networks to produce content code and style code, respectively. Also, we save the pre-calculated deep feature map $e_{I_A}$ for $I_A$. Compiling these two feature maps together following the principle of Eq.~\ref{eq1}, we have the composed feature map:
\begin{equation}
\label{feature_composition}
\hat{e}_g^{(i,j)} =\left\{
\begin{aligned}
& e_{I_g}^{(i,j)} , & M_e^{(i,j)}=1, \\
& e_{I_A}^{(i,j)}, & M_e^{(i,j)}=0.
\end{aligned}
\right.
\end{equation} Note, $M_e$ is the down-sampled version of $M$ to suit for the spatial size of $\hat{e}_g$. After feature composition, $\hat{e}_g$ is fed to two sub-networks ($H_c$ and $H_s$) to generate the content code $C_g$ and style code $S_g$, respectively:
\begin{equation}
\label{eq_subnetwork}
C_g = H_c(\hat{e}_g), \quad S_g = H_s(\hat{e}_g).
\end{equation}
These two codes are then fed to the generator to produce $I_h$. 

Though $I_h$ is composed by $I_g$ and $I_A$, it is not necessarily ensured that $I_h$ should look similar to $I_g$ in the masked area while other parts are similar to $I_A$. To enforce such constraint, we additionally apply a {\bf C}ontent {\bf A}lignment loss, which we term as $\mathcal{L}_{\mathrm{CA}}$:
\begin{small}
\begin{equation}
    \label{eq_hybrid}
    \mathcal{L}_{\mathrm{CA}} = \mathcal{L}_1 (I_h \cdot M, I_g \cdot M) +  \mathcal{L}_1 (I_h \cdot (1 - M), I_A \cdot (1 - M)).
\end{equation}
\end{small} 
We adopt $\mathcal{L}_1$ norm for this loss, however, we believe any loss function that ensures content consistency can be applied here. Similar to the training of Eq.~\ref{eq_self_supervision}, $\mathcal{L}_{\mathrm{CA}}$ is also alternatively trained once with an interval of $K$ iterations.

There are couple of works that also devote to region-wise image editing, such as SEAN~\cite{SEAN} and SMIS~\cite{SMIS}. However, these works require semantic labels in both training and inference while our method is devoid of this requirement. This gives our model a larger prospect of real world applications where semantic labels are not obtainable. During training, we produce randomly generated masks as~\cite{PartialConv,FreeForm} to assist the calculation of Eq.~\ref{eq_hybrid}. For inference, the mask is provided by the users as interested areas for stylization.

\section{Experiments}

\begin{figure*}[htbp]
	\centering
	\includegraphics[width=\textwidth]{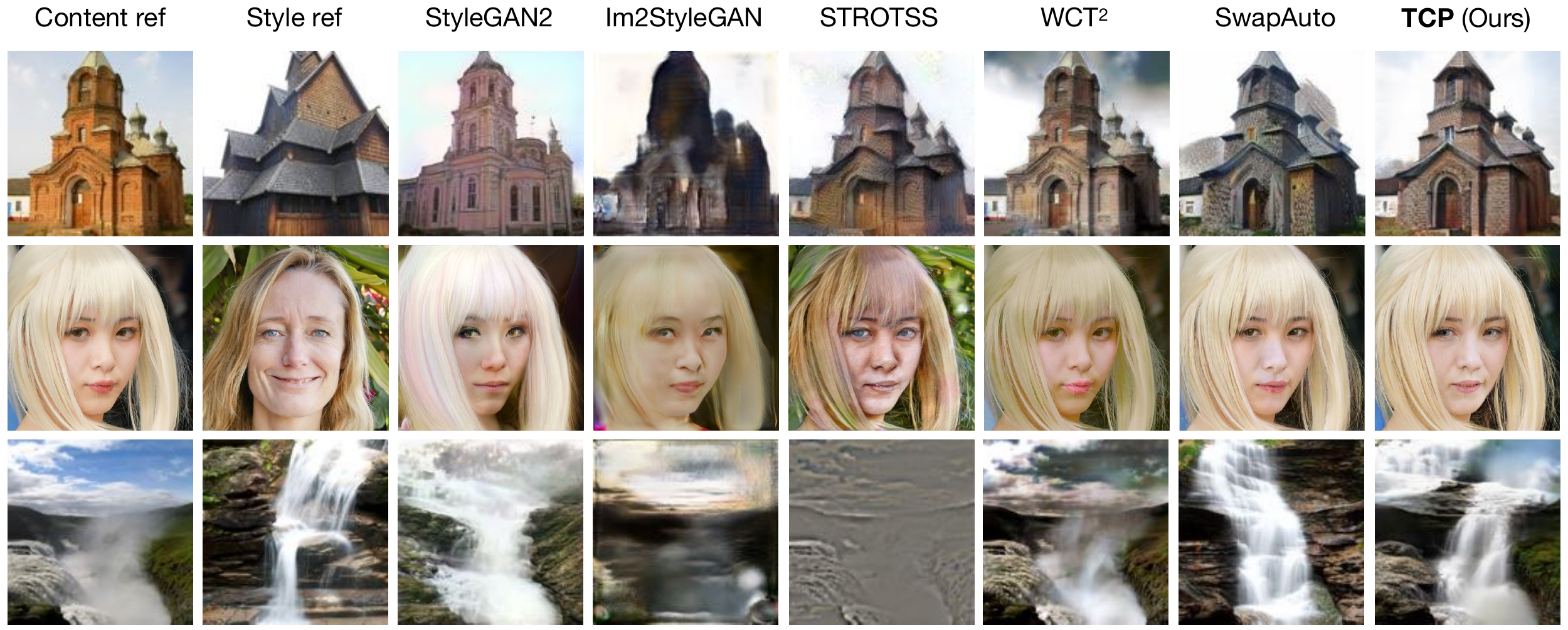}
	\caption{Qualitative comparison of global image transfer on various datasets. Upper: LSUN-Church example; Middle: FFHQ; Lower: Waterfall.
	\label{fig:comparison_on_global_image_transfer}
	}
\end{figure*}

\subsection{Experiment settings}


\vspace{0.5em} \noindent {\bf Datsets.}~Our model is trained on LSUN-Church~\cite{LSUN}, Flickr Faces HQ(FFHQ)~\cite{StyleGAN}, CelebAMask-HQ~\cite{CELEBA, CELEBAHQ, CELEBAMASKHQ}, Mountains, Waterfalls, Disney faces and Ukiyoe faces datasets. There are about 126K and 70K images for LSUN-Church and FFHQ, respectively. CelebAMask-HQ contains 30K CelebAHQ~\cite{CELEBAHQ} face images with 19 different region categories and we use this dataset to evaluate the region-wise control ability of different methods. We follow SwapAuto~\cite{SwapAuto} to collect approximately 58K mountain images and 100K waterfall images from the Flickr image website\footnote{\url{https://www.flickr.com/}} as the Mountains and Waterfalls datasets. Disney faces and Ukiyoe faces datasets both contain around 20K images, which are collected via~\cite{Disney}. Besides, we have trained on FFHQ dataset together with Disney faces or Ukiyoe faces dataset for the evaluation of cross-domain style transfer.

\vspace{0.5em} \noindent {\bf Metrics.}~We use Fr\'{e}chet Inception Distance (FID)~\cite{FID} to calculate the distance between the distributions of synthesized images and the distribution of real images. Lower FID generally hints better quality or fidelity of the generated images. Besides, as indicated in \cite{SwapAuto}, we use Single Image FID~\cite{SinGAN} and Self-similarity Distance \cite{self-similarity} to evaluate the style change and content change of the hybrid images with regard to the input content images, respectively.

\vspace{0.5em} \noindent {\bf Baselines.}~Our most direct baseline is SwapAuto~\cite{SwapAuto} because our self-supervision constraints are employed on the original architecture of SwapAuto. Since the official code of SwapAuto is not released, we re-implement their method based on the descriptions in their paper. We also follow SwapAuto to compare with StyleGAN2~\cite{STYLEGAN2}, Im2StyleGAN~\cite{Im2StyleGAN}, STROTSS~\cite{STROTSS}, WCT$^2$~\cite{WCT2} about the global style transfer results. Besides, to evaluate the performance of the region-wise editing ability of our model, we compare with SEAN~\cite{SEAN}, which supports region-wise lable-to-image translation and style manipulation.

\subsection{Comparisons}

\begin{table}[htbp]
	\centering
		\resizebox{0.48\textwidth}{!}{
		\begin{tabular}{|l|c|c|c|c|c|c|} 
			\hline
			Models & SSIM $\uparrow$ & RMSE $\downarrow$ & PSNR $\uparrow$ & FID $\downarrow$ & SSD $\downarrow$ & SIFID $\uparrow$ \\ \hline
			SEAN~\cite{SEAN} & 0.607 & 10.16 & 27.99 & 29.71 & 0.081 & 1.293 \\
			SwapAuto~\cite{SwapAuto} & 0.608 & 10.02 & 28.12 & 20.05  & 0.089 & 1.216  \\ \hline
			\textbf{TCP} (Ours) & 0.770 & 9.67 & 28.43 & 21.72 & 0.064 & 1.127 \\  
			\hline
	\end{tabular}
	}
	\caption{Region-wise stylization quantitative comparison on the CelebAMask-HQ testing set selected by SEAN. We use {\bf TCP} to denote our result. 
	\label{tab:regionwise_stylization}
	}
\end{table}

\begin{table}[tb]
	\centering
    	\resizebox{0.48\textwidth}{!}{
    		\begin{tabular}{|l|c|c|c|c|c|c|c|c} 
    			\hline
    			\multirow{2}{*}{Method}&
    			\multicolumn{3}{c|}{FFHQ}&
    			\multicolumn{3}{c|}{LSUN-Church} \\
    			\cline{2-4} \cline{5-7} 
    			 & FID$\downarrow$ & SSD $\downarrow$  & SIFID$\uparrow$ & FID$\downarrow$ & SSD $\downarrow$  & SIFID$\uparrow$ \\ \hline
    			StyleGAN2~\cite{STYLEGAN2} & 77.74 & 0.070 & 3.452 & 57.92 & 0.093 & 1.676 \\
    			Im2styleGAN\cite{Im2StyleGAN} & 115.16 & 0.082 & 6.036 & 221.45 & 0.099 & 2.262\\
    			STROTSS~\cite{STROTSS} & 100.29 & 0.052 & 4.257 & 69.85 & 0.051 & 2.120 \\
    			WCT$^2$~\cite{WCT2} & 37.65 & 0.031 & 3.207 & 34.88 & 0.037 & 1.383 \\
    			SwapAuto~\cite{SwapAuto} & 58.57 & 0.052 & 1.971 & 52.79 & 0.079 & 1.720 \\
    			\hline
    			\textbf{TCP} (Ours) & 51.30 & 0.042 & 4.450 & 50.41 & 0.081 & 1.807 \\ \hline
    			
    		\end{tabular}
    	}
	\caption{Quantitative comparison of global image transfer on the FFHQ~\cite{StyleGAN} and LSUN-Church~\cite{LSUN} datasets. Note the unit for SIFID is $10^{-5}$.
	\label{tab:ablation_baseline}
	}
\end{table}

\begin{figure*}[htbp]
	\centering
	\includegraphics[width=\textwidth]{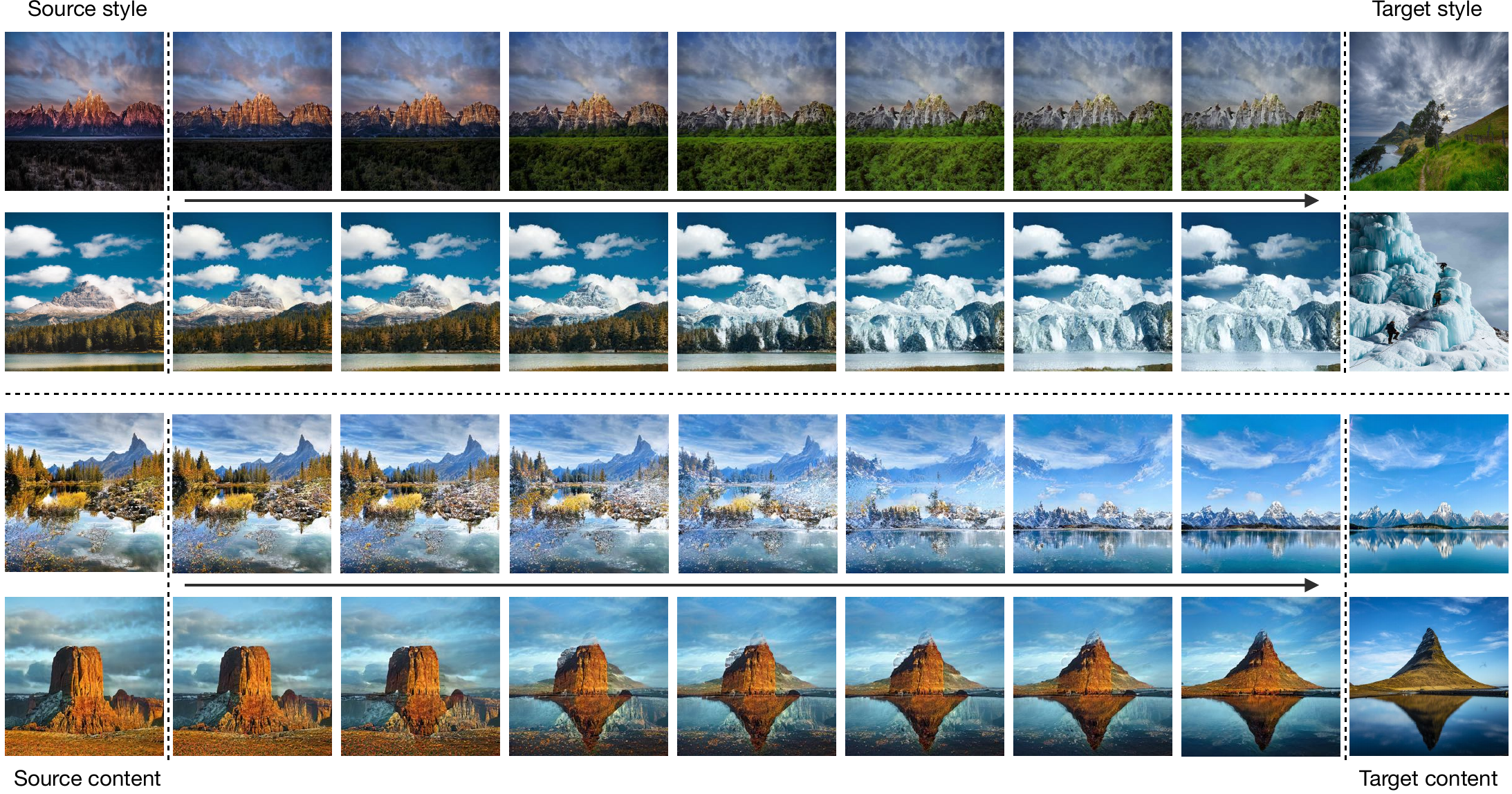}
	\caption{Code interpolation results between two real images. The left-most images are the source images and the right-most images are the style/content reference images. By varying the proportion of style/content codes, we can produce a gradual change from source to target. The reference images are taken from \url{https://pixabay.com/}.
	\label{fig:code_interpolation}
	}
\end{figure*}

\begin{figure}[htbp]
	\centering
	\includegraphics[width=0.48\textwidth]{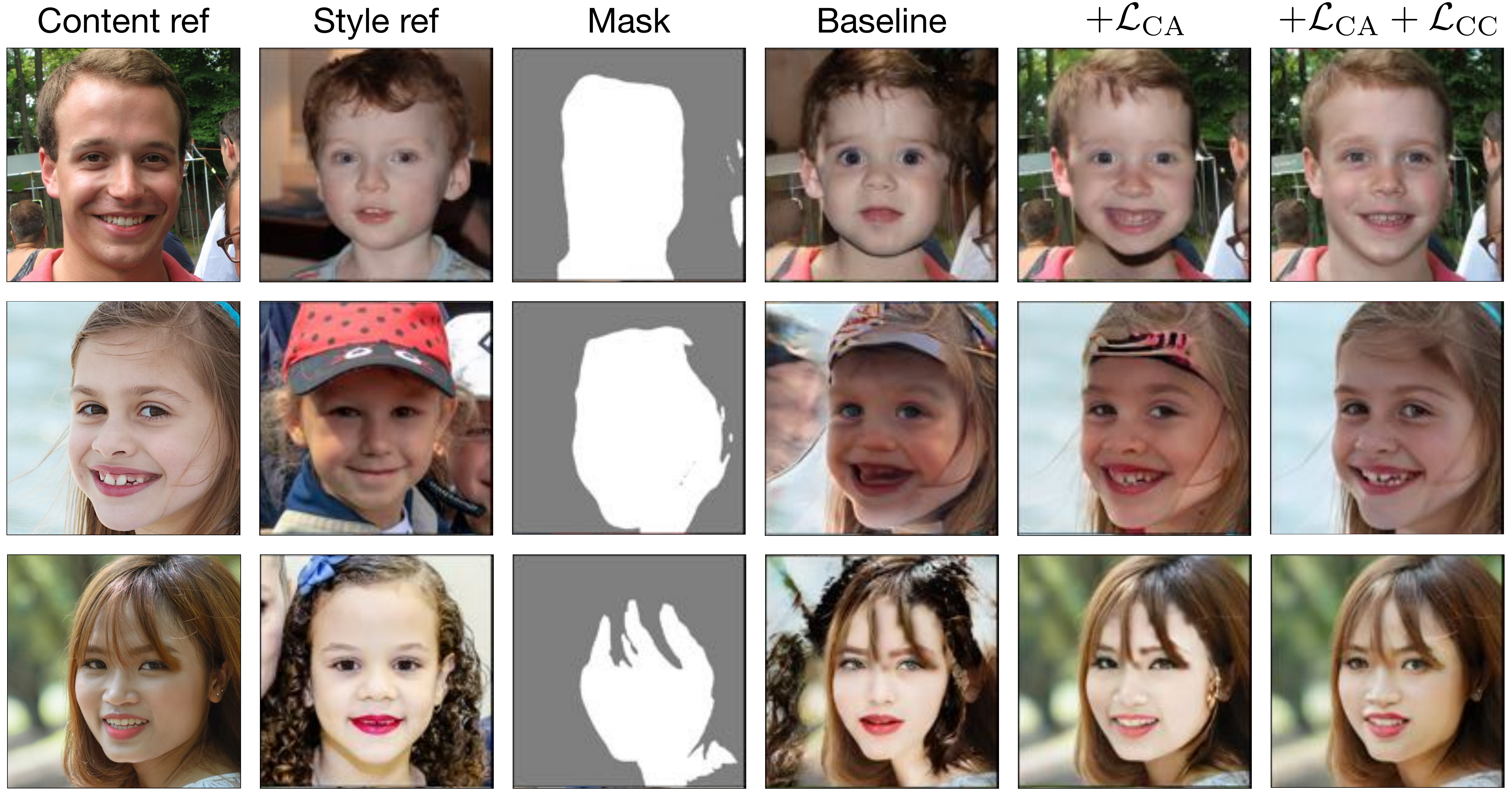}
	\caption{Ablation qualitative comparisons on FFHQ.
	\label{fig:ablation}
	}
\end{figure}

\subsubsection{Region-wise stylization}

In Fig.~\ref{fig:comparison_on_regionwise_stylization}, we show the region-wise stylization comparison with SEAN~\cite{SEAN} and SwapAuto~\cite{SwapAuto} on the CelebAMask-HQ dataset~\cite{CELEBA,CELEBAHQ,CELEBAMASKHQ}. To compare with SwapAuto~\cite{SwapAuto} in terms of region-wise control, we perform the structure code manipulation as described in their paper, which is similar to our feature composition in Eq.~\ref{feature_composition}. Note since the code of SwapAuto is not released, we run a open-sourced code\footnote{\url{https://github.com/rosinality/swapping-autoencoder-pytorch}} using the default setting as described in their paper. 
SEAN performs better than our model in transferring styles, but it loses some structure details when the semantic label is not precise, \eg, the shoulder of the girl in the left figure is not reconstructed. Our method shows the best content/structure consistency to the input content reference. SwapAuto rarely maintains the structure because its style code and content code are entangled somehow that the style code also has some structural information. Since we enforce our code disentanglement loss during training, our method is not troubled with this issue.

We follow most of the metric comparison (SSIM, RMSE, PSNR) in SEAN~\cite{SEAN} to show the image transfer performance of different methods on the CelebAMask-HQ dataset~\cite{CELEBA, CELEBAHQ, CELEBAMASKHQ}. We download their train/test splits, containing 28K/2K images separately. For each test image, the corresponding style image is randomly chosen to generate the stylized image. All the quantitative results are showed in Table \ref{tab:regionwise_stylization}. We recommend to see these numerical results in a comprehensive manner rather than treating them separately because single metric may bias our judgement. For example, very high SSIM means the results may be too similar to the inputs and there is little style change.
Generally, we wise the results to stick to the content reference (larger SSIM, smaller RMSE and SSD) but their styles to be changed (larger SIFID). Besides, the results should be realistic-looking (smaller FID and larger PSNR). Using the recommended perspective to see the results, our method is comparatively better than the other two methods.

\subsubsection{Global image transfer}

Besides the comparison on region-wise stylization, we also compare the globally image transfer performance with previous methods, specifically, StyleGAN2~\cite{STYLEGAN2}, Im2StyleGAN~\cite{Im2StyleGAN}, STROTSS~\cite{STROTSS}, WCT$^2$~\cite{WCT2} and SwapAuto~\cite{SwapAuto}. For fair comparison, we download the results from the server of SwapAuto\footnote{\url{http://efrosgans.eecs.berkeley.edu/SwappingAutoencoder/}}, which contains the comparison results and testing pairs of all mentioned methods on FFHQ, LSUN-Church and Flickr Waterfall datasets, respectively. 
We use the same testing pairs and images to test our model.
The visual comparison is shown in Fig.~\ref{fig:comparison_on_global_image_transfer}. Generally, WCT$^2$ shows the best structure consistency with the content reference, but the styles are not perfectly reflected. This conclusion is also mentioned in~\cite{SwapAuto}. Our results keep structurally consistent to the content references while SwapAuto seems to lose structure consistency, especially at the waterfall example. Our method also shows the superiority to extract tiny appearance features, \eg, in the second row, our result exhibits the correct eye color of the style reference.

For quantitative comparison, our method achieves lower FID, lower SSD and higher SIFID than most of previous methods on FFHQ and LSUN-Church datasets, showed in Table~\ref{tab:ablation_baseline}. As mentioned earlier, lower FID and SSD means the more plausible results, while higher SIFID means more style changes. WCT$^2$ has the lowest FID and SSD, but the lowest SIFID indicates that the generated images might be close to the source image without much style change. The FID scores of other methods except WCT$^2$ are lower than our method by a large margin. The results clearly show the superiority of our method over previous methods.

\begin{figure}[htbp]
	\centering
	\includegraphics[width=0.48\textwidth]{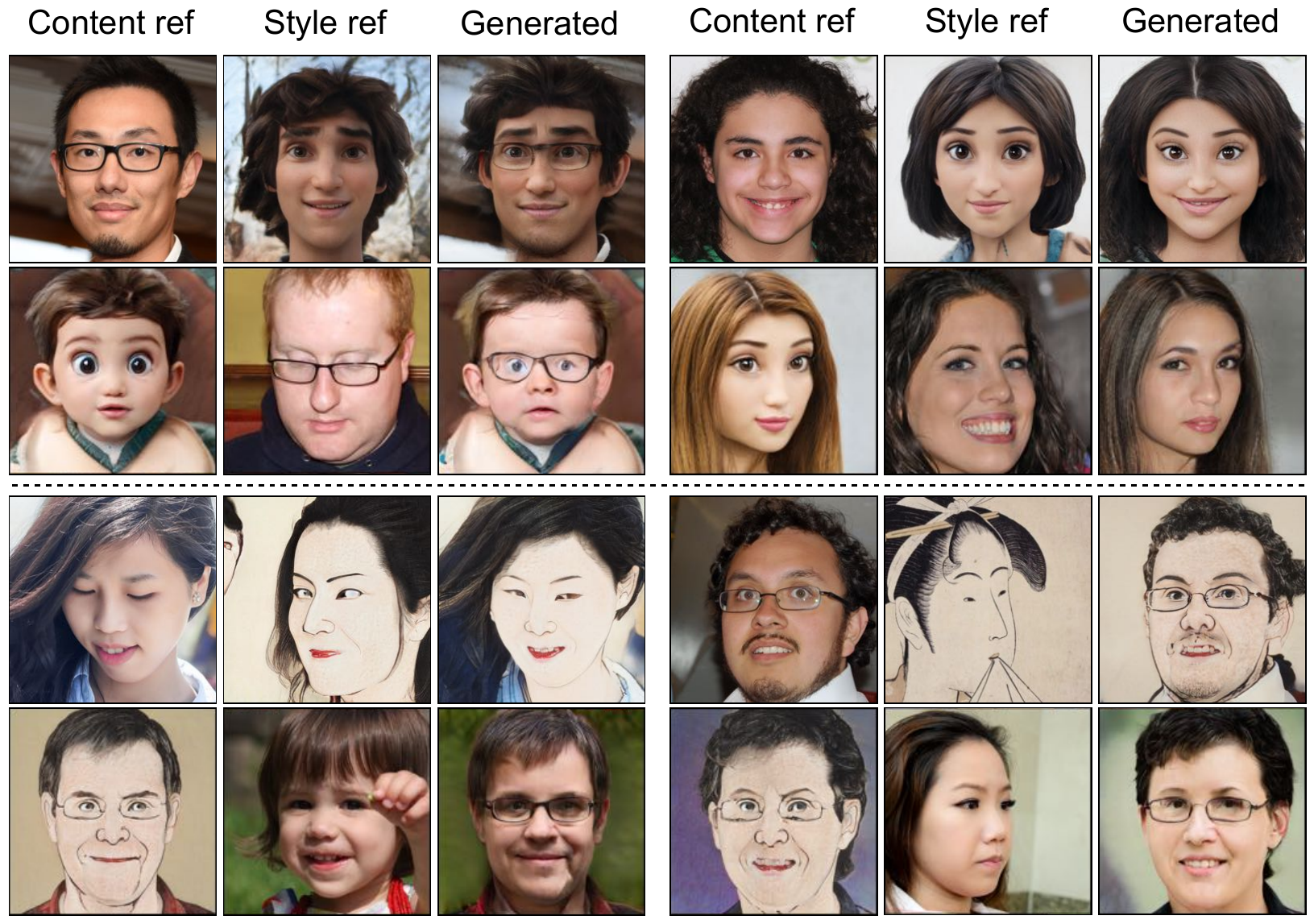}
	\caption{Cross-domain image transfer results of training on combined datasets (top two rows: FFHQ-Disney faces; bottom two rows: FFHQ-Ukiyoe faces). 
	\label{fig:cross_domain}
	}
\end{figure}

\subsection{Ablation study}

The main contributions of this work are: 1) the content alignment loss $\mathcal{L}_{\mathrm{CA}}$ for region-wise stylization 2) the code consistency loss $\mathcal{L}_{\mathrm{CC}}$ for content and style code disentanglement. In Fig.~\ref{fig:ablation}, we successively add these two losses to the baseline model (SwapAuto) to show the effects of these two loss terms. All these experiments are conducted on FFHQ\cite{StyleGAN} with image size downsampled to 512. The semantic masks used for region-wise control are obtained by using a face parsing model.\footnote{\url{https://github.com/zllrunning/face-parsing.PyTorch}}

From the figure, when adding $\mathcal{L}_{\mathrm{CA}}$ to the Baseline model, given a face mask representing the foreground area to be stylized, the $+\mathcal{L}_{\mathrm{CA}}$ model is able to keep the background unchanged while the Baseline model lacks such ability. This clearly shows that $\mathcal{L}_{\mathrm{CA}}$ loss is able to make the model learn a disentanglement between foreground and background, thus enabling region-wise stylization. But both Baseline and $+\mathcal{L}_{\mathrm{CA}}$ models have distorted faces and tend to ``copy'' some elements in the style reference image. This phenomenon implies the structure controllability is falsely captured by the style code. However, our full pipeline ($+\mathcal{L}_{\mathrm{CA}} + \mathcal{L}_{\mathrm{CC}}$) 
does not suffer from this issue, because we make a clearer code disentanglement to the content codes and style codes. The above comparisons clearly show the effectiveness of both our techniques.

\subsection{Applications}

\vspace{0.5em} \noindent {\bf Code interpolation.}~Since our encoder projects images to editable style and content codes, we show in Fig.~\ref{fig:code_interpolation} that our model is able to generate a sequence of images given a source content/style and a target content/style by linearly interpolating between the content/style codes. These results illustrate a gradual change from the source to the target and the generated images are also meaningful, especially for style interpolation.

\vspace{0.5em} \noindent {\bf Cross-domain image style transfer.}~
In this section, we show that our model is able to perform cross-domain image style transfer by directly training on a dataset whose data is from the combination of two domains. Suppose we have $\mathcal{D}_A$ domain representing real faces, and $\mathcal{D}_B$ domain for disney characters, our model is able to perform either in-domain image transfer ($\mathcal{D}_A\rightarrow \mathcal{D}_A$ and $\mathcal{D}_B\rightarrow \mathcal{D}_B$, results in supplementary materials due to page limit) or cross-domain image transfer ($\mathcal{D}_A\rightarrow \mathcal{D}_B$ and $\mathcal{D}_B\rightarrow \mathcal{D}_A$). Some qualitative results of cross-domain transfer are demonstrated in Fig.~\ref{fig:cross_domain}. 
From the figure, though the performances are not totally symmetric for each pair of the supported cross-domain transfer, the experiment shows the possibility of image transfer of arbitrary domains if they share similar underlying structure patterns and are trained in a combined dataset. We may no longer need to train a single model for each style transfer direction. There are more details and more results in the supplementary materials.

\section{Conclusion}

In this paper, we build an auto-encoder that aims at photorealistic region-wise style editing on real images. We propose two loss functions based on self-supervision to modulate the training process. The code alignment loss constrains our model to conduct an explicit style and content code disentanglement. The content consistency loss enables the model to perform region-wise style editing without using other annotations like semantic labels and object locations, \etc. Extensive experiments show the superiority of our method over previous SOTAs. We believe our method has a broad prospect in real image editing, animation creation, virtual reality, \etc. In the future, we plan to step further on more flexible and explicit form of code disentanglement strategy and explore more interesting applications using our model.


\section*{Acknowledgements}
We would like to thank all those who provided help with this work. 
We would also like to thank the anonymous reviewers for their helpful suggestions. The work is supported by The ShanghaiTech University for the project: Shanghai YangFan Program(No. 21YF1429500).

\bibliographystyle{ACM-Reference-Format}
\balance
\bibliography{egbib}

\end{document}